%% file: main.tex
\documentclass[10pt,twocolumn,letterpaper]{article}

\usepackage{cvpr}              
\usepackage{float}
\usepackage{mathrsfs}
\usepackage{bm}
\usepackage{multirow}
\usepackage[accsupp]{axessibility} 

\input{preamble}

\definecolor{cvprblue}{rgb}{0.21,0.49,0.74}
\usepackage[pagebackref,breaklinks,colorlinks,citecolor=cvprblue]{hyperref}

\def\paperID{806} 
\def\confName{CVPR}
\def\confYear{2024}

\title{GaussianAvatar: Towards Realistic Human Avatar Modeling \\ from a Single Video via Animatable 3D Gaussians}

\author{
Liangxiao Hu$^{\dag, 1}$, Hongwen Zhang$^2$, Yuxiang Zhang$^3$, Boyao Zhou$^3$, Boning Liu$^3$,\\ Shengping Zhang$^{*, 1, 4}$, Liqiang Nie$^1$\\
$^1$Harbin Institute of Technology $^2$Beijing Normal University \\
$^3$Tsinghua University $^4$Peng Cheng Laboratory\\
\tt\small \{lx.hu, s.zhang\}@hit.edu.cn, zhanghongwen@bnu.edu.cn, yx-z19@mails.tsinghua.edu.cn \\
\tt\small \{bzhou22, liuboning\}@mail.tsinghua.edu.cn, nieliqiang@gmail.com
}

\begin{document}

\twocolumn[{%
\maketitle
\begin{figure}[H]
\hsize=\textwidth %
\centering
\includegraphics[width=\textwidth]{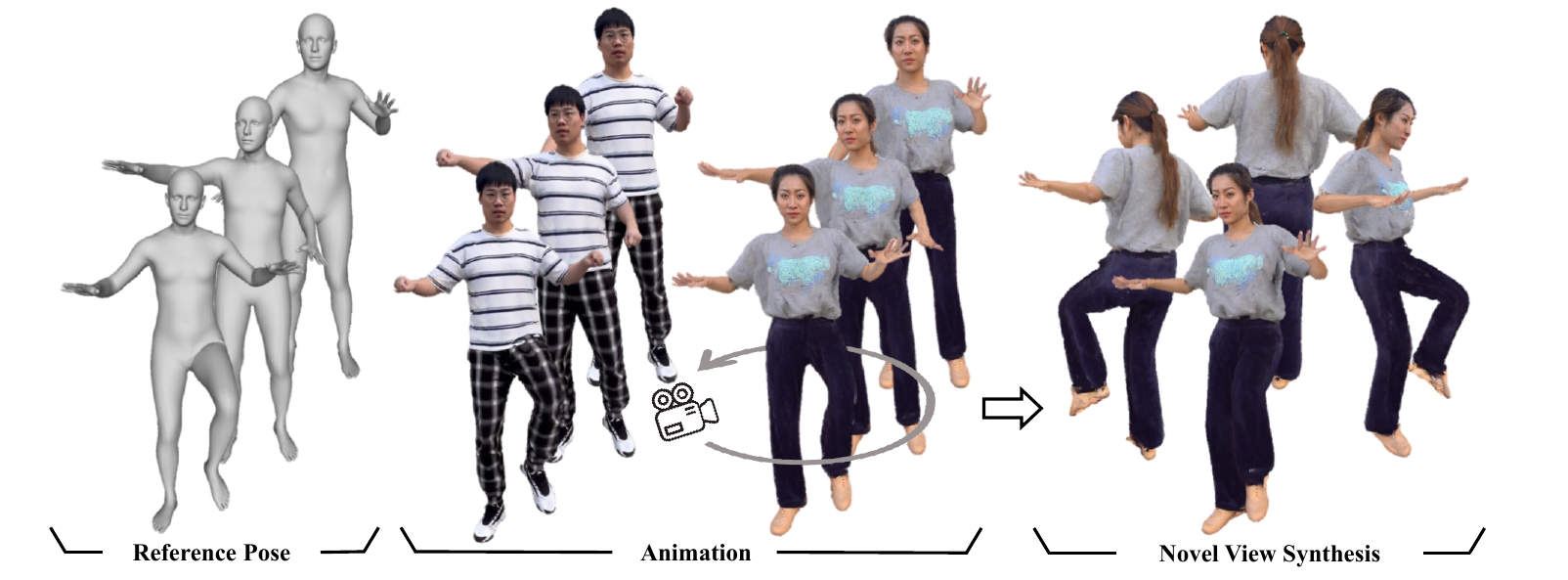}
\caption{We propose GaussianAvatar, which learns animatable 3D Gaussians to represent detailed human avatars from a single video. Our method maintains a 3D consistent appearance even when animated by out-of-distribution motions.} 
\label{fig:teaser}
\end{figure}
}]

\let\thefootnote\relax\footnotetext{$^\dag$ Work done during an internship at Tsinghua University.}
\let\thefootnote\relax\footnotetext{$^*$ Corresponding author.}

\maketitle

\input{sections/0_abstract}

\input{sections/1_introduction}

\input{sections/2_related}
\input{sections/3_1_method}
\input{sections/3_2_method}

\input{sections/3_3_method}
\input{sections/3_4_method}

\input{sections/4_experiments}
\input{sections/5_conclusion}
\clearpage
{
    \small
    \bibliographystyle{ieeenat_fullname}
    \bibliography{main}
}
\input{sections/supp_no_re}


\end{document}

%% file: preamble.tex
\usepackage[dvipsnames]{xcolor}


%% file: sections/0_abstract.tex
\begin{abstract}
We present GaussianAvatar, an efficient approach to creating realistic human avatars with dynamic 3D appearances from a single video.
We start by introducing animatable 3D Gaussians to explicitly represent humans in various poses and clothing styles.
Such an explicit and animatable representation can fuse 3D appearances more efficiently and consistently from 2D observations.
Our representation is further augmented with dynamic properties to support pose-dependent appearance modeling, where a dynamic appearance network along with an optimizable feature tensor is designed to learn the motion-to-appearance mapping.
Moreover, by leveraging the differentiable motion condition, our method enables a joint optimization of motions and appearances during avatar modeling, which helps to tackle the long-standing issue of inaccurate motion estimation in monocular settings.
The efficacy of GaussianAvatar is validated on both the public dataset and our collected dataset, demonstrating its superior performances in terms of appearance quality and rendering efficiency.
%
%
%
%
%
%
%
The code and dataset are available at {\small\url{https://github.com/aipixel/GaussianAvatar}}.

\end{abstract}

%% file: sections/1_introduction.tex
\section{Introduction}
\label{sec:introduction}
Creating a customized human avatar from a single video has great potential for many applications including virtual and augmented reality, the Metaverse, gaming, and movie industries.
This task is appealing yet challenging, as the monocular observations are highly underdetermined for the modeling of a 3D animatable avatar.
Meanwhile, the inaccurate body motion estimations and the complex wrinkle deformations also make it extremely difficult to create a realistic avatar from a single video.
%
%
%

%
%
The modeling of 3D human avatars from monocular videos involves a fusion process of 2D observations to a 3D consistent human model.
For this purpose, existing solutions have leveraged both implicit and explicit representations to create the base model of avatars.
Among them, implicit methods~\cite{jiang2022neuman, weng2022humannerf, guo2023vid2avatar, yu2023monohuman, jiang2023instantavatar} define a deformable human NeRF to fuse the image observation from current motion space to a canonical space by inverse skinning.
%
However, the implicit 3D volume is inefficient in representing human surfaces and the inverse skinning also introduces ambiguous correspondences during the fusion process.
These issues make it hard for implicit solutions to capture fine-grained details of moving people.
%
As the avatar appearances are concentrated around human surfaces, explicit representations are much more efficient in modeling avatars.
Previous attempts~\cite{huang2023efficient, zhao2022high} have employed differentiable mesh rendering to reconstruct the human surface, but these methods struggle to preserve wrinkle details due to a fixed mesh topology. 
%
%
On the other hand, point-based representations~\cite{ruckert2022adop, wiles2020synsin, lassner2021pulsar, yifan2019differentiable,zheng2023pointavatar} are more effective to handle flexible topology but require millions of points to capture detailed appearances.
How to represent humans remains one of the fundamental problems for avatar modeling.
%
%

%

When creating an animatable avatar from a video of moving people, the algorithm is required to learn the relationships between body motions and corresponding appearances. 
However, the motions estimated from monocular videos are typically faulty, leading to large artifacts in the modeling of dynamic cloth deformations.
%
To address this issue, previous works~\cite{weng2022humannerf, jiang2022neuman, su2022danbo, alldieck2018video, te2022neural, jiang2023instantavatar, yu2023monohuman} have attempted to optimize body motions along with the learning of animatable avatar volumes.
%
%
%
As the body motions are explicitly represented as parametric meshes, the implicit 3D volume of previous methods makes the optimization indirect and less effective.
This issue has even become the main obstacle to achieving high-quality avatar modeling from monocular videos.

%
%
%

To tackle the above issues, we introduce new representations and solutions to achieve high-quality avatar modeling from a single video.
The key insight of our solution is to model dynamic human surfaces explicitly and optimize both the motion and appearances jointly in an end-to-end manner.
To this end, we propose GaussianAvatar, a method to reconstruct human avatars with dynamic appearances using the 3D Gaussian representation~\cite{kerbl20233dgaussian}.
As an explicit representation, 3D Gaussian can be easily reposed from the canonical space to the motion space via a forward skinning process.
Such an animatable 3D Gaussian representation bypasses the inverse skinning process used in the aforementioned NeRF-based methods and overcomes the previous one-to-many issue~\cite{chen2021snarf} during canonicalization.
Based on the animatable 3D Gaussian, our method can fuse 3D appearances more consistently from 2D observations to a canonical 3D space.
%
To model dynamic human appearances under different poses, we additionally add pose-dependent properties to 3D Gaussians and incorporate them with the canonical human surfaces.
Inspired by previous work~\cite{ma2021power, vakalopoulou2018atlasnet}, we learn a dynamic appearance network on the 2D manifolds of the underlying human mesh to predict dynamic properties of 3D Gaussians. 
However, due to the strong bias of limited training poses, modeling dynamic appearance solely conditioned on pose information struggles to generalize to novel views and poses.
To address this issue, we introduce an optimizable feature tensor to capture a coarse global appearance of human avatars.
Subsequently, we incorporate pose-dependent effects on the feature tensor to decode fine-grained details such as wrinkles.
%

%

As the proposed animatable 3D Gaussians are differentiable with respect to the motion conditions, it enables a joint optimization of motion and appearances.
This merit allows our network to refine the motion along with the avatar modeling process, which helps to tackle the long-standing issue of inaccurate motion estimation in monocular settings.
Besides, the refined motion further enhances the accuracy of avatar modeling since the 3D appearance fusion process relies on motion-based skinning.
As shown in our experiments, our method is quite robust to initial motion estimation and has the ability to correct the misalignment of motion capture results.
%
%
%
%
%
%

\medskip
\noindent
 
To summarize, our main contributions are as follows:

\begin{itemize}
    \item We introduce animatable 3D Gaussians for realistic human avatar modeling from a single video. By representing human surfaces explicitly, our method can fuse 3D appearances more consistently and efficiently from 2D observations.

    \item We augment the animatable 3D Gaussians with dynamic properties to support pose-dependent appearance modeling, where a dynamic appearance network along with an optimizable feature tensor is designed to learn the motion-to-appearance mapping.

    \item We propose to jointly optimize the motion and appearance during the avatar modeling, enabling our method to correct the misalignment of initial motion and improve the final appearance quality.


\end{itemize}

%% file: sections/2_related.tex
\section{Related Work}
\paragraph{Neural Rendering for Human Reconstruction.}
Without the need to define a template mesh for avatar modeling~\cite{xu2011video, bagautdinov2021driving, habermann2021real}, neural rendering has emerged as a potent technique that enables learning avatars directly from images.
Here we briefly review precious work that aims to reconstruct humans using neural rendering.
%

Due to the high-quality rendering of neural radiance field~\cite{mildenhall2020nerf}, various efforts~\cite{liu2021neural, peng2021neural, peng2022animatable, li2022tava, wang2022arah, zheng2022structured, li2023posevocab, zheng2023avatarrex} have been made to reconstruct the dynamic appearance of moving people.
Neural Body~\cite{peng2021neural} associates a latent code to each SMPL~\cite{loper2015smpl} vertex to encode the appearance, which is transformed into observation space based on the human pose.
Neural Actor~\cite{liu2021neural} learns a deformable radiance field with SMPL as guidance and utilizes a texture map to improve its final rendering quality.
TAVA~\cite{li2022tava} proposes to jointly model the non-rigid warping field and shading effects directly conditioned on the pose vectors.
Posevocab~\cite{li2023posevocab} designs joint-structured pose embeddings to encode the dynamic appearances under different key poses, such embeddings can better learn joint-related appearance.
NeRF-based methods have demonstrated appealing rendering results on human avatar reconstruction, but still struggle to represent human surfaces with the implicit 3D volume.
Explicit modeling of human surfaces is a more straightforward way for this task, as the dynamic appearance of humans is mostly reflected on the human surface.
%
%

Explicit representations have great potential for human reconstruction.
In HF-Avatar~\cite{zhao2022high}, meshes are used as the base representation in a coarse-to-fine framework that combines neural texture with dynamic surface deformation for avatar creation.
EMA~\cite{huang2023efficient} proposes Meshy neural fields to reconstruct human avatars by optimizing the canonical mesh, material, and motion dynamics through inverse rendering in an end-to-end process.
PointAvatar~\cite{zheng2023pointavatar} employs a deformable point-based representation to separate source color into intrinsic albedo and normal-dependent shading.
DVA ~\cite{remelli2022drivable} extends mixtures of volumetric primitives~\cite{Lombardi21} for human avatar modeling.
All these attempts have demonstrated the significant potential of explicit representations and their under-exploration.
However, meshes are constrained by fixed topologies, whereas point clouds demand a multitude of points to encompass intricate details.
%
%
%
3D Gaussians~\cite{kerbl20233dgaussian} have showcased their capability in various human tasks~\cite{li2023animatablegaussians,xu2023gaussianheadavatar,zheng2023gpsgaussian,shao2023control4d}.
%
%
In essence, 3D Gaussians hold promise for human avatar reconstruction and are currently
a subject of active research.

\paragraph{Avatar Modeling from Monocular Videos.}
Numerous methods investigate reconstructing humans from single images or monocular videos.
Regression-based methods\cite{saito2019pifu, saito2020pifuhd, zheng2020pamir, xiu2021icon, xiu2023econ, huang2020arch, he2021arch++, he2020geo} directly recover clothed 3D humans just from a single image.
While these methods produce attractive results, they can not recover the dynamic appearance in the reconstruction across the entire sequence.
Traditional methods aim to capture human dynamics by tracking individuals in videos using pre-scanned rigged templates~\cite{deepcap, habermann2019livecap, MonoPerfCap}.
However, the pre-scanning and manual rigging processes prevent their real-life applications.
\cite{alldieck2018video, guo2021humanperformance, Moon_2022_ECCV_ClothWild} try to bypass the requirement for predefined human models but face challenges in preserving fine details because of the fixed mesh resolution.
The emergence of neural radiance fields~\cite{mildenhall2020nerf} has facilitated the creation of various techniques for reconstructing animatable avatars from monocular videos~\cite{weng2022humannerf, jiang2022selfrecon, jiang2022neuman, su2022danbo, su2021a-nerf, chen2021animatable, su2023npc}.
%
%
However, inaccuracies in estimating human motions from monocular videos result in pronounced artifacts.
To address the problem, HumanNeRF~\cite{weng2022humannerf} solves for an update to the inaccurate poses.
NeuMan~\cite{jiang2022neuman} introduces an error-correction network to enable training with erroneous estimates.
Vid2Avatar\cite{guo2023vid2avatar} and InstantAvatar~\cite{jiang2023instantavatar} also jointly optimize motions by back-propagating the gradient of the image reconstruction loss to the pose parameters.
MonoHuman~\cite{yu2023monohuman} introduces bi-directional constraints to alleviate ambiguous correspondence on novel poses.
%
Attempting to address the inaccurate pose estimation issue with implicit human NeRF models has proven to be inefficient and imprecise.
Consequently, we endeavor to address this issue by leveraging an explicit 3D Gaussian representation.

%% file: sections/3_1_method.tex
\begin{figure*}[]
\centering
\includegraphics[width=\textwidth,]{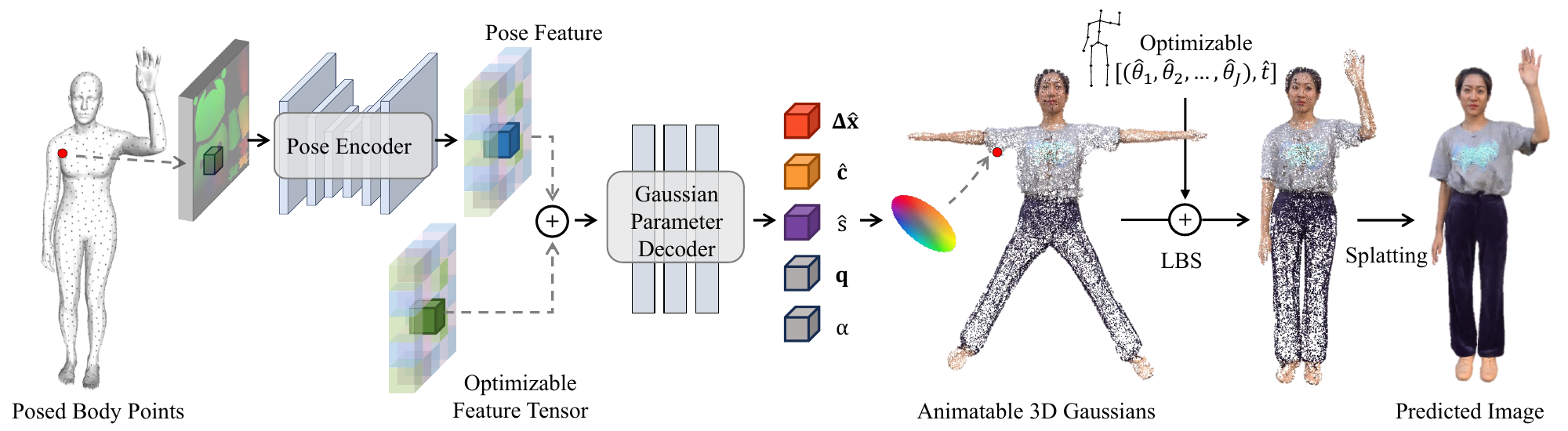}
\caption{\textbf{Overview of GaussianAvatar. }
Given a fitted SMPL or SMPL-X model on the current frame, we sample the points on its surface and record their positions on a UV positional map $I$, which is then passed to a pose encoder to obtain the pose feature.
An optimizable feature tensor is pixel-aligned with the pose feature and learned to capture the coarse appearance of humans.
Then the two aligned feature tensors are input into the Gaussian parameter decoder, which predicts each point's offset $\Delta \hat{ \mathbf{x}}$, color $\hat{ \mathbf{c}}$, and scale $\hat{s}$.
These predictions, along with the fixed rotations $\mathbf{q}$ and opacity $\alpha$, collectively constitute the animatable 3D Gaussians in canonical space.
}
\vspace{-0.3cm}
\label{fig:overview}
\end{figure*}

\section{Method}
\label{sec:method}
Our goal is to create a human avatar that enables free-viewpoint rendering and realistic animation using a single video.
We first introduce an expressive representation, namely animatable 3D Gaussians, to represent human avatars in Sec.~\ref{sec:representation}.
With this representation, modeling the dynamic appearance of humans can be regarded as dynamic 3D Gaussian property estimation.
Then we build a dynamic appearance network along with an optimizable feature tensor to learn the motion-to-appearance mapping in Sec.~\ref{sec:dynamicmodeling}.
%
%
%
To mitigate the artifacts caused by inaccurate pose estimation, we adopt a joint motion and appearance optimization approach for refining human poses during training (Sec.~\ref{sec:pose}).
%
%
In the following, we provide a comprehensive explanation of the technical details.

%% file: sections/3_2_method.tex
\subsection{Animatable 3D Gaussians}
\label{sec:representation}
Point-based representation~\cite{ma2021power, lin2022learning, zhang2023closet, prokudin2023dynamic, zhou2020reconstructing} has proven its topological flexibility in generating realistic human avatars from scans.
Extending this explicit representation to create human avatars from images is a significant endeavor.
With this goal in mind, we introduce a novel representation, termed animatable 3D Gaussians, which effectively reconstructs human surfaces with a 3D consistent appearance.
%
%

3D Gaussian Splatting~\cite{kerbl20233dgaussian} is a point-based scene representation that allows high-quality real-time rendering.
The scene representation is parameterized by a set of static 3D Gaussians, each of which has the following parameters: 3D center position $\mathbf{x} \in \mathbb{R}^3$, color $\mathbf{c} \in \mathbb{R}^3$, opacity $\alpha \in \mathbb{R}$, 3D rotation in form of quaternion $\mathbf{q} \in \mathbb{R}^4$ and 3D scaling factor $\mathbf{s} \in \mathbb{R}^3$.
With these properties, we can generate rendered images from any viewpoint.
We refer readers to ~\cite{luiten2023dynamic, kerbl20233dgaussian} for the rendering details of 3D Gaussians.
%

%
%
To extend this representation for human avatar modeling, we integrate it with either 
 the SMPL~\cite{loper2015smpl} or SMPL-X~\cite{SMPL-X:2019} model as follows:
\begin{equation}
\label{eq:avatar}
\begin{split}
G(\bm{\beta}, \bm{\theta}, \mathbf{D}, \mathbf{P}) &= Splatting(W(\mathbf{D}, J(\bm{\beta}), \bm{\theta}, \omega), \mathbf{P}), 
\end{split}
\end{equation}
where $G(\cdot)$ represents a rendered image, and $Splatting(\cdot)$ denotes the rendering process of 3D Gaussians from any viewpoint, $W(\cdot)$ is a standard linear blend skinning function employed for reposing 3D Gaussians, $\mathbf{D} = T(\bm{\beta}) + dT $ represents the locations of 3D Gaussians in canonical space, formed by adding corrective point displacements $ dT $ on the template mesh surface $ T(\bm{\beta})$, $\mathbf{P}$ denotes the remaining properties of 3D Gaussians, excluding the positions.
$\bm{\beta}$ and $\bm{\theta}$ are the shape and pose parameters, $J(\bm{\beta})$ outputs 3D joint locations.
Note that we propagate the skinning weight $\omega$ from the vertices of the SMPL or SMPL-X model to the nearest 3D Gaussians.
With the proposed representation, we can now repose these canonical 3D Gaussians to the motion space for free-view rendering.


%

\subsection{Dynamic 3D Gaussian Property Estimation}
\label{sec:dynamicmodeling}
Following the proposed animatable Gaussians, human appearances are determined by the point displacements $dT$ and properties $\mathbf{P}$.
Modeling dynamic human appearances can be regarded as estimating these dynamic properties.
To model dynamic human appearances under various poses, we introduce a dynamic appearance network along with an optimizable feature tensor to predict these pose-dependent properties of 3D Gaussians.
Despite sharing similarities in network structure with~\cite{ma2021power}, we present this framework for a distinct purpose.
In~\cite{ma2021power}, the feature tensor serves to decouple the pose-independent human shape from the decoder, and we repurpose it to capture a coarse global appearance of human avatars.
The motivation behind this modification is that directly learning a mapping from human poses to dynamic properties is susceptible to overfitting on the limited training poses.
To integrate the global appearance into the feature tensor, we introduce a two-stage training strategy, as discussed in Sec.~\ref{sec:losses}.
\begin{figure}[t]
\includegraphics[width=\linewidth]{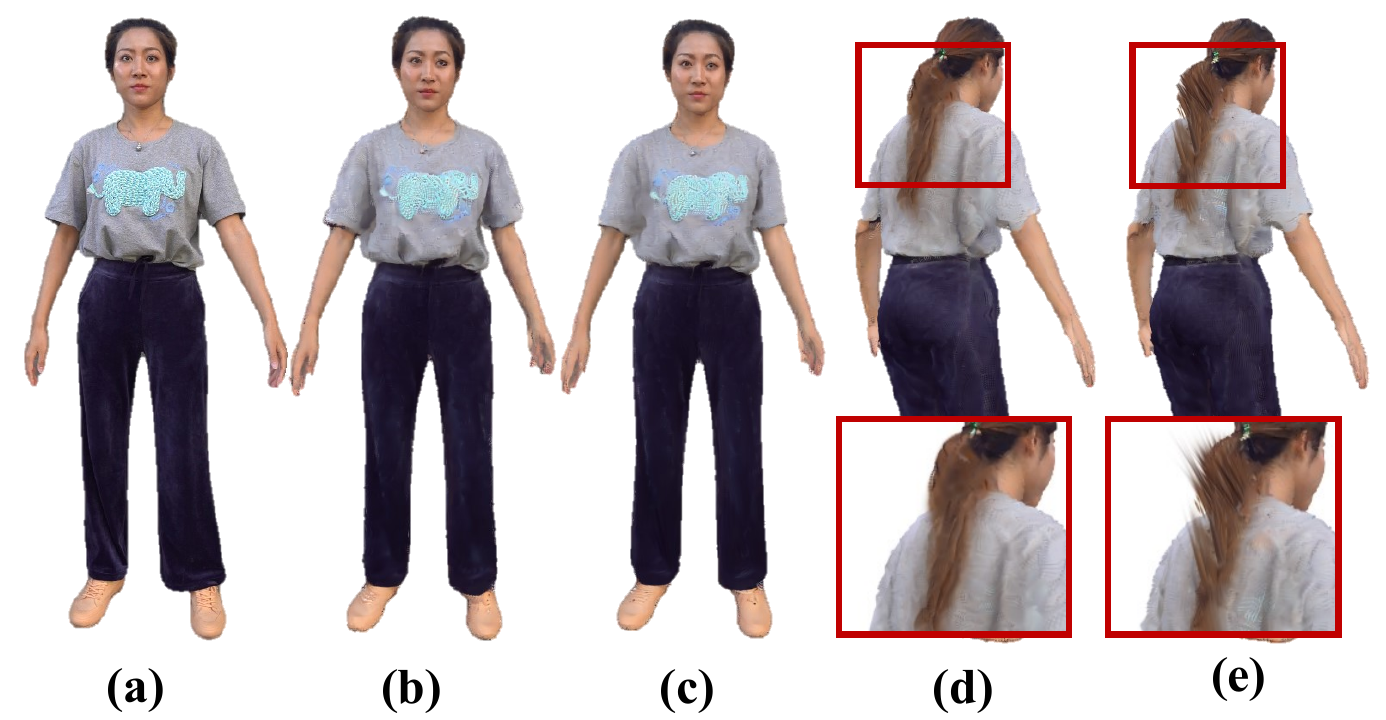}
\caption{\textbf{Effect of iostropy of 3D Gaussians.}
(a) Input image, (b)(d) front and back views trained with isotropic 3D Gaussians, (c)(e) front and back views trained with anisotropic 3D Gaussians.
}
\vspace{-0.3cm}
\label{fig:abiso}
\end{figure}

The dynamic appearance network is designed to learn a mapping from a 2D manifold representing the underlying human shape to the dynamic properties of 3D Gaussians as follows:
\begin{equation}
\label{formula:gaussian}
    f_\phi: \mathcal{S}^2 \in \mathbb{R}^3 \rightarrow \mathbb{R}^7.
\end{equation}
As shown in Fig.~\ref{fig:overview}, the 2D human manifold $\mathcal{S}^2$ is depicted by a UV positional map $I \in R^{H \times W \times 3}$, where each valid pixel stores the position $(x, y, z)$ of one point on the posed body surface.
The final predictions consist of per point offset $\Delta \hat{ \mathbf{x}} \in \mathbb{R}^3$, color $\hat{ \mathbf{c}} \in \mathbb{R}^3$, and scale $\hat{s} \in \mathbb{R}$ on the canonical surface.
%

%
%
Instead of predicting all properties, we make these slight adjustments to our task based on experimental results.
%
%
Due to the unbalanced viewpoints in monocular videos, anisotropic 3D Gaussians are prone to learning an inaccurate 3D shape to fit the most frequently seen view, resulting in poorer performance in side views.
As shown in Fig.~\ref{fig:abiso}, training our network with isotropic 3D Gaussians yields superior results.
We ensure isotropy among all 3D Gaussians by maintaining uniform size across dimensions.
Therefore, we introduce a scaler $\hat{s}$ to represent the scales of 3D Gaussians and set rotations $ \mathbf{q}$ as $[1, 0, 0, 0]$.
Through experimental observations, we notice that the network tends to learn an opacity value ($\alpha$) of zero on the boundary to correct the human shape.
To address this, we fix opacity $\alpha = 1$ to keep all 3D Gaussians visible, enforcing the network to predict accurate positions of the 3D Gaussians.
%

%

%

%

%
%
The dynamic appearance network consists of two parts: a pose encoder and a Gaussian parameter decoder.
The pose encoder takes the UV positional map of posed body points as input to generate a pose-conditioned feature tensor $O \in \mathbb{R}^{H \times W \times C}$.
We then integrate the pixel-aligned optimized feature tensor $F \in \mathbb{R}^{H \times W \times C}$ with the pose features before feeding it into the Gaussian parameter decoder to generate final predictions.
Following this, we add the predicted offsets to the canonical 3D Gaussians and associate the predicted properties with the corresponding 3D Gaussians.
With the estimated poses, we can then repose the canonical 3D Gaussians to the motion space for rendering.
%
%
%

%

%

%

%

%

%

%

%

%% file: sections/3_3_method.tex
\subsection{Joint Motion and Appearance Optimization}
\label{sec:pose}
However, owing to the imprecise estimation of human poses $\bm{\theta} = ({\theta}_1, {\theta}_2, ..., {\theta}_J)$ and translations $\mathbf{t}$ from monocular videos, the reposed 3D Gaussians in motion space are inaccurately represented and may lead to unsatisfactory rendering outcomes.
To address this, we propose to jointly optimize human motions and appearances.
To optimize human motions with image loss, we solve for an update $(\Delta \bm{\theta}, \Delta \mathbf{t})$ to the estimated body poses and translations as follows:
\begin{equation}
\label{eq:update}
     \hat{\bm{\Theta}} = (\bm{\theta} + \Delta \bm{\theta}, \mathbf{t} + \Delta \mathbf{t}).
\end{equation}
We modify $\bm{\theta}$ in Eq.~\ref{eq:avatar} using $\hat{\bm{\Theta}}$ to render the proposed animatable 3D Gaussians differentiable with respect to the motion conditions.
%
%
Different from previous work~\cite{weng2022humannerf, guo2023vid2avatar, jiang2022neuman} that jointly optimizes human poses via inverse skinning, we optimize the updates in a forward skinning process, which benefits both motion and appearance optimization.

%% file: sections/3_4_method.tex
\subsection{Training Strategy}
\label{sec:losses}
In this section, we outline our approach to training the network with inaccurate human motions.
We conduct a two-stage optimization process using different loss functions.
In the first stage, we aim to fuse the sequential appearances to the optimizable feature tensor and conduct motion optimization to get accurate poses for the dynamic appearance network.
In this stage, we optimize the framework without incorporating any pose-dependent information by excluding the training of the pose encoder.
%
Specifically, we utilize the following loss functions to train our network:
\begin{equation}
\label{formula:loss_stage1}
\begin{split}
       \mathcal{L}_{stage_1} = &\lambda_{rbg}\mathcal{L}_{rbg} + \lambda_{ssim}\mathcal{L}_{ssim} + \lambda_{lpips}\mathcal{L}_{lpips}\\
        + &\lambda_{f}\mathcal{L}_{f} + \lambda_{offset}\mathcal{L}_{offset} + \lambda_{scale}\mathcal{L}_{scale},
\end{split}
\end{equation}
where $\mathcal{L}_{rbg}$, $\mathcal{L}_{ssim}$, and $\mathcal{L}_{lpips}$ are the L1 loss,  SSIM loss~\cite{wang2004image}, and LPIPS loss~\cite{zhang2018unreasonable}, respectively.
$\mathcal{L}_{f}$, $\mathcal{L}_{offset}$, $\mathcal{L}_{scale}$ calculate the L2-norm of the feature map, predicted offsets and scales, respectively.
We set $\lambda_{rbg}=0.8$, $\lambda_{ssim}=0.2$, $\lambda_{lpips}=0.2$, $\lambda_{f}=1$, $\lambda_{offset}=10$, $\lambda_{scale}=1$. 
After the first stage of training, we obtain more accurate human motions and an optimized feature tensor $F$.
The optimized feature tensor $F$ captures a coarse appearance of human avatars.
In the second stage, we incorporate the pose features encoded by the pose encoder with the trained feature tensor $F$.
We replace $\mathcal{L}_{f}$ with the L2-norm loss $\mathcal{L}_{p}$, which plays the same role as $\mathcal{L}_{f}$  in regularizing the limited pose space.
By penalizing the pose-dependent features, we can eliminate the strong bias of limited training poses and thus generalize to unseen viewpoints and poses.
%

%

%

%

%

%

%

%

%

%

%

%

%

%

%

%

%

%

%

%

%

%

%% file: sections/4_experiments.tex
\input{tables.tex}
\tablesnapshot
\tableneuman
\tabledyn
\section{Experiments}
\label{sec:experiments}
\subsection{Datasets and Metrics}
\label{sec:dataset}
\noindent \textbf{People-Snapshot Dataset.} This dataset~\cite{alldieck2018video} comprises videos of individuals rotating in front of a stationary camera.
To ensure a fair quantitative comparison, we follow the same evaluation protocol outlined in InstantAvatar~\cite{jiang2023instantavatar}.
%
%

\noindent \textbf{NeuMan Dataset.} To assess even more challenging scenarios, we employ outdoor collections from the NeuMan dataset~\cite{jiang2022neuman}.
These videos are recorded using a mobile phone for moving individuals. 
Specifically, we select four sequences (bike, citron, seattle, jogging) that exhibit most body regions and contain minimal blurry images.
We initialize the estimated poses with an off-the-shelf method~\cite{sun2021monocular}, which is also utilized in \cite{jiang2023instantavatar}.
\noindent \textbf{DynVideo Dataset.} With the limited cloth deformation presented in the above two datasets, we propose the Dynvideo dataset to capture dynamic human appearance.
To this end, we utilize a mobile phone to record videos of a character performing various movements, especially rotation, in front of the device.
Each of these videos takes about one minute and provides a comprehensive and detailed representation of human movement.
We also provide the corresponding SMPL parameter sequences for all videos, obtained by our proposed method.
This dataset serves as a valuable resource for evaluating reconstruction quality, with a particular emphasis on dynamic appearances.
\noindent \textbf{Evaluation Metrics.} We consider three metrics: PSNR, SSIM~\cite{wang2004image}, and LPIPS~\cite{zhang2018unreasonable} to access the reconstruction quality on three datasets.
%

%

\begin{figure}[t]
\includegraphics[width=\linewidth]{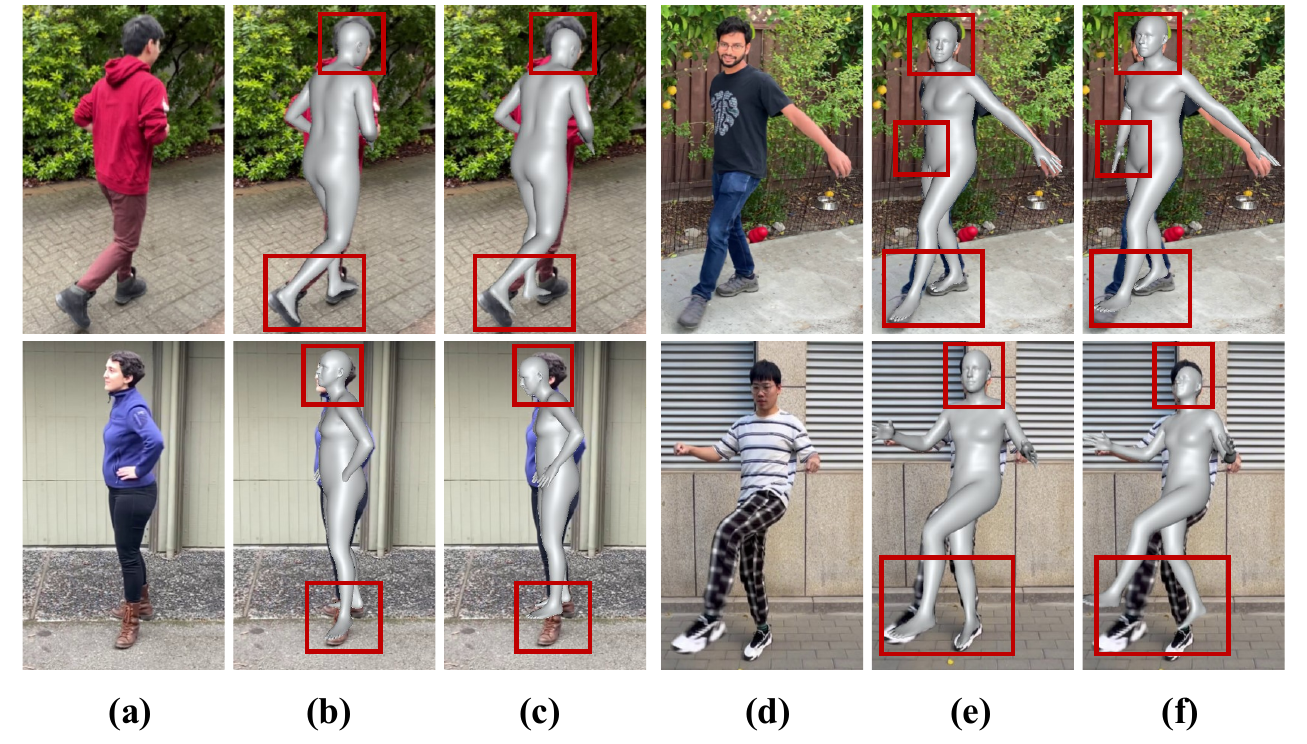}

\caption{\textbf{Motion optimization results.}
(a)(d) Original image, (b)(e) our optimized SMPL, (c)(f) ROMP~\cite{sun2021monocular} estimates.
}
\label{fig:abpose}
\end{figure}

\begin{figure}[t]
\includegraphics[width=\linewidth]{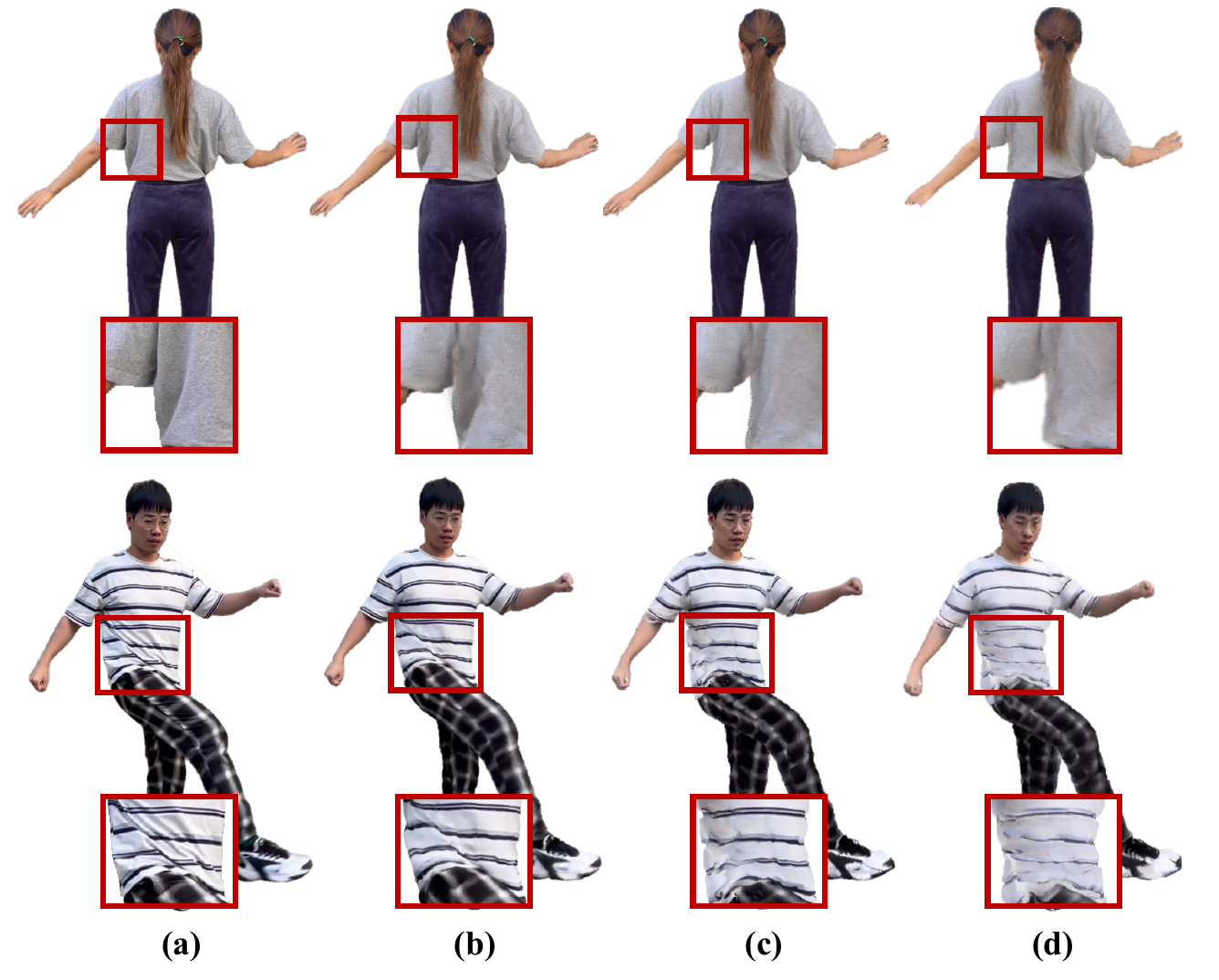}

\caption{\textbf{Qualitative ablation studies.} (a) Ground truth, (b) baseline + Opt. + Dyn., (c) baseline + Opt., (d) baseline.}
\vspace{-0.3cm}
\label{fig:abc2f}
\end{figure}

\subsection{Experimental Settings}
\label{sec:setting}
\noindent \textbf{Implementation details.}  
We employ a U-Net~\cite{ronneberger2015u} for extracting pose-dependent features, and the Gaussian parameter decoder is implemented as an 8-layer multilayer perceptron (MLP).
We sample approximately 200,000 points on the SMPL mesh surface.
The entire framework is trained on a single NVIDIA RTX 3090 GPU, with training times ranging from 0.5 to 6 hours.

\noindent \textbf{Baseline.} 
To showcase the efficacy of the proposed modules, our method can be partitioned into three components: baseline, motion optimization, and dynamic appearance modeling.
As our baseline, we suspend the motion optimization process and exclude the pose-dependent information from the pose encoder.

\begin{figure*}[]
\centering
\includegraphics[width=\textwidth,]{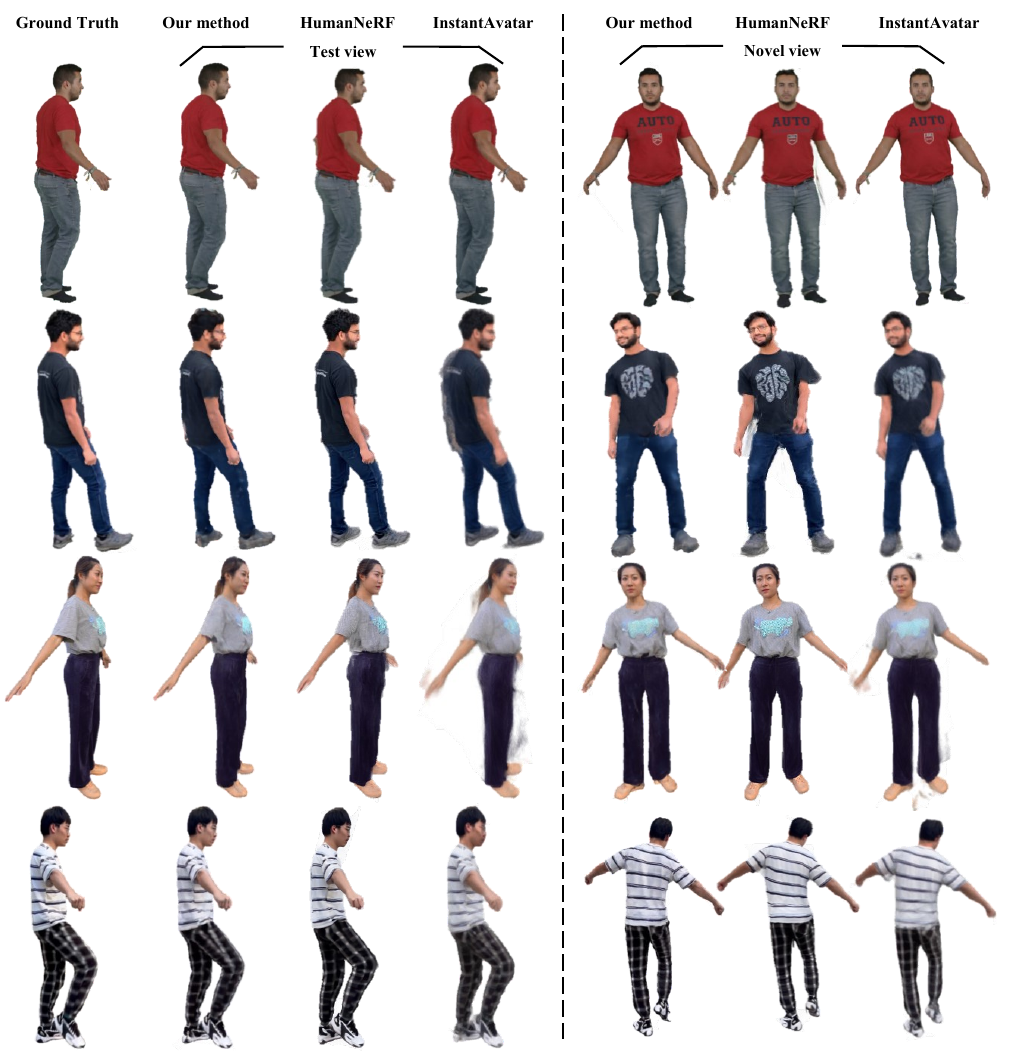}
\caption{\textbf{Qualitative comparison of novel view synthesis.} We compare the novel view synthesis quality on the People-Snapshot dataset (first row), NeuMan dataset (second row), and DynVideo dataset (last two rows).}
\vspace{-0.1cm}
\label{fig:novelview}
\end{figure*}

\begin{figure*}[]
\centering
\includegraphics[width=\textwidth,]{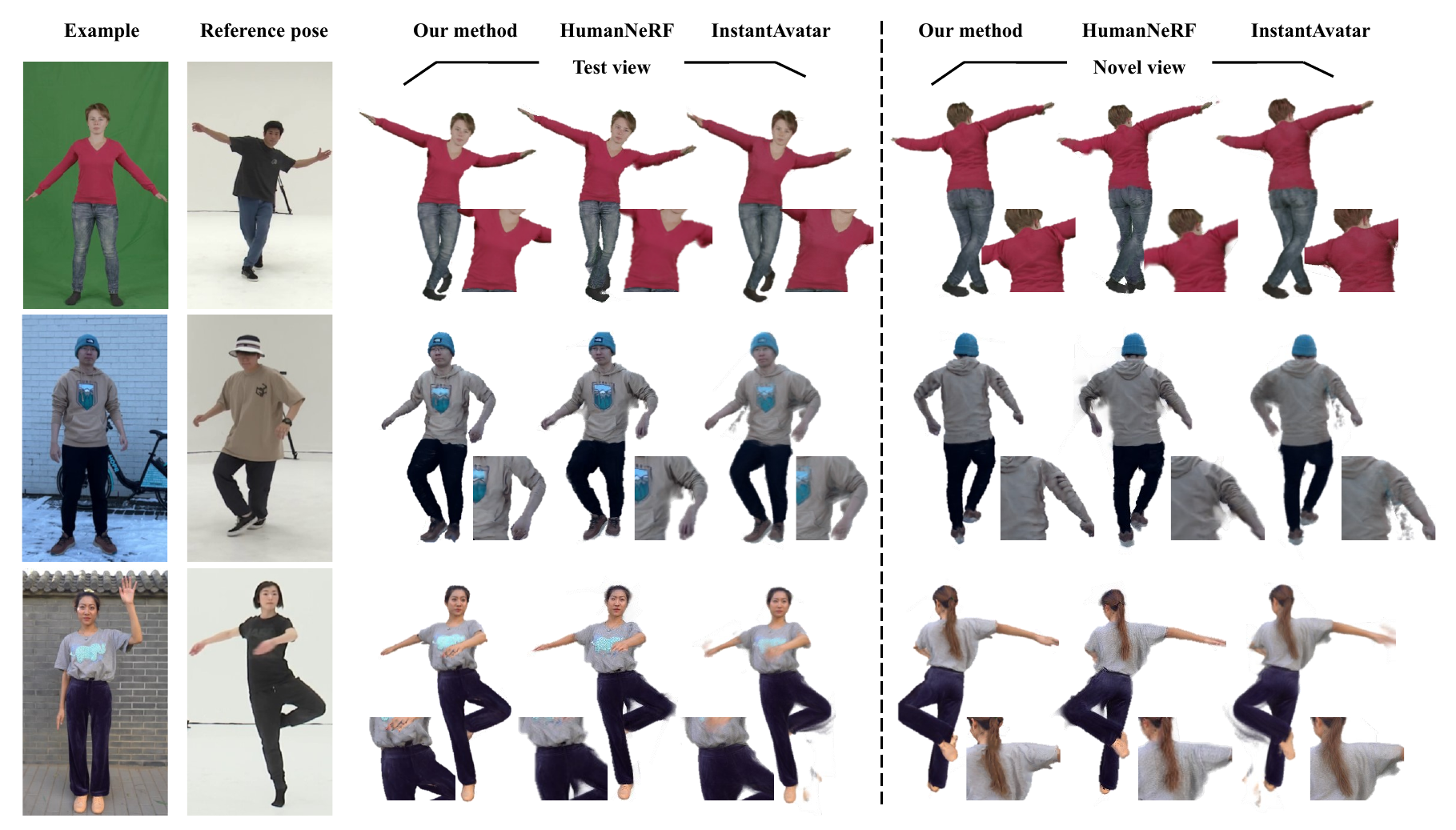}
\caption{\textbf{Animation results on out-of-distribution motions.} We compare the animation results on the People-Snapshot dataset (first row), NeuMan dataset (second row), and DynVideo dataset (last row).}
\vspace{-0.1cm}
\label{fig:novelpose}
\end{figure*}
\noindent \textbf{Methods for Comparison.} We compare our method against (1) HumanNeRF~\cite{weng2022humannerf}, which implicitly represents a human avatar with a canonical appearance neural field and a motion field; (2) InstantAvatar~\cite{jiang2023instantavatar}, which achieves fast avatar modeling by using several acceleration strategies for neural fields.

\subsection{Comparisons with the State of the Art}
\label{sec:comparison}
We report the numeric evaluation results on three datasets.
As shown in Table~\ref{tab:peoplesnapshot},~\ref{tab:neuman}, and~\ref{tab:dynvideo}, our proposed method outperforms all baselines on all metrics for recovering more details of dynamic appearance and correcting the artifacts caused by initialized poses. 
%
%
To demonstrate the qualitative evaluation of these datasets, we also visualize the novel view synthesis results on the test splits.
As implicit representations, HumanNeRF and InstantAvatar are prone to generate ghosting effects at boundary areas, as shown in Fig.~\ref{fig:novelview}.
%
%
InstantAvatar lacks the capability to model pose-dependent deformations, leading to challenges in effectively handling the dynamic appearance of moving people.
%

To showcase the robustness of our avatar modeling, we collect more challenging poses~\cite{li2021learn} captured by the monocular camera and evaluate our method and other methods for avatar animation and novel view synthesis.
%
As ground truth is unavailable for these out-of-distribution poses, we illustrate qualitative results of novel pose and view synthesis in Fig.~\ref{fig:teaser} and Fig.~\ref{fig:novelpose}. 
Our method generates realistic animation results with these challenging poses, demonstrating a consistent 3D appearance in novel views with respect to other methods.
\subsection{Ablation Studies}
\label{sec:ablation}
In this section, we conduct ablation studies to validate each component of our methods.
As shown in Table~\ref{tab:peoplesnapshot},~\ref{tab:neuman}, and~\ref{tab:dynvideo}, our proposed motion optimization module dramatically improves over the baseline on all metrics, demonstrating its effectiveness in modeling human avatars.
To highlight the effectiveness of our method in human motion optimization, we illustrate the initialized poses obtained by ROMP~\cite{sun2021monocular} and the optimized ones by our method on NeuMan and DynVideo datasets in Fig.~\ref{fig:abpose}.
Experiments demonstrate that joint motion optimization is capable of correcting inaccurate motion estimation, even for the side and back views.
Furthermore, our approach readily extends to enhance the accuracy of existing motion capture methods~\cite{pymaf2021, pymafx2023, zhang2023realtime}.
We also notice that the dynamic appearance modeling achieves superior results on NeuMan and DynVideo datasets in Table~\ref{tab:neuman} and \ref{tab:dynvideo} and slightly favorable outcomes in Table~\ref{tab:peoplesnapshot}, considering that People-Snapshot has limited pose variation with respect to the other two datasets.
%
%
In Fig.~\ref{fig:abc2f}, we further illustrate the visible improvements of each component of our method.
The motion optimization scheme improves the global quality of the baseline for better pose estimation.
The dynamic appearance modeling additionally preserves pose-dependent details such as cloth wrinkles on the human surface.

%% file: tables.tex
\newcommand{\tablesnapshot}{
\begin{table*}[t]
\centering
\definecolor{Gray}{gray}{0.85}
\resizebox{\textwidth}{!}{
\begin{tabular}{l|ccc|ccc|ccc|ccc}
\toprule
\multicolumn{1}{c}{\multirow{2}[4]{*}{Method}} & \multicolumn{3}{c}{male-3-casual} & \multicolumn{3}{c}{male-4-casual} & \multicolumn{3}{c}{female-3-casual} & \multicolumn{3}{c}{female-4-casual}\\
\cmidrule{2-13} \multicolumn{1}{c}{} & PSNR$\uparrow$  & SSIM$\uparrow$  & \multicolumn{1}{c}{LPIPS$\downarrow$}  & PSNR$\uparrow$  & SSIM$\uparrow$  &\multicolumn{1}{c}{LPIPS$\downarrow$}  & PSNR$\uparrow$  & SSIM$\uparrow$  & \multicolumn{1}{c}{LPIPS$\downarrow$}  & PSNR$\uparrow$  & SSIM$\uparrow$  & LPIPS$\downarrow$ \\

\midrule
HumanNeRF~\cite{weng2022humannerf} & 26.90 & 0.9605 & 0.0181 & 25.50 & 0.9397 & 0.0357& 24.46 & 0.9516 & 0.0269& 27.07 & 0.9615 & 0.0152 \\
InstantAvatar \cite{jiang2023instantavatar}  & 29.53 & 0.9716 & 0.0155 & 27.67 & 0.9626 & 0.0307 & 27.66 & 0.9709 & 0.0210 & 29.11 & 0.9683 & 0.0167 \\

\midrule
Baseline & 27.71 & 0.9713 & 0.0218 & 25.09 & 0.9614 & 0.0306 & 24.80 & 0.9609 & 0.0310 & 26.16 & 0.9610 & 0.0227 \\
Baseline + Opt. & \textbf{30.98} & \textbf{0.9791} & \textbf{0.0143} & 28.77 & 0.9753 & 0.0230  & \textbf{29.59} & \textbf{0.9768} & \textbf{0.0220} & 30.83 & 0.9768 & 0.0145  \\
Baseline + Opt. + Dyn. & \textbf{30.98} & 0.9790 & 0.0145 & \textbf{28.78} & \textbf{0.9755} & \textbf{0.0228}  & 29.55 & 0.9762 & 0.0225 & \textbf{30.84} & \textbf{0.9771} & \textbf{0.0140}  \\

\bottomrule

\end{tabular}
}
\caption{ \textbf{Quantitative evaluation on the People-Snapshot~\cite{alldieck2018video} dataset.} Opt. denotes the motion optimization and Dyn. refers to dynamic appearance modeling.}
\vspace{-0.2cm}
\label{tab:peoplesnapshot}
\end{table*}
}

\newcommand{\tableneuman}{
\begin{table}[t]
\small
\centering
\begin{tabular}{lccc}

\hline  Method & PSNR $\uparrow $ &  SSIM $\uparrow$  & LPIPS $\downarrow $\\ 
\hline
HumanNeRF~\cite{weng2022humannerf}  & 27.06 & 0.9669 & 0.0192 \\ 
InstantAvatar~\cite{jiang2023instantavatar}  & 28.47  &  0.9715 &  0.0277\\ 

\hline
  Baseline & 27.06 & 0.9692 & 0.0183  \\
  Baseline + Opt. & 29.93  & 0.9794 & 0.0126   \\
  Baseline + Opt. + Dyn. & \textbf{29.94}  & \textbf{0.9795} & \textbf{0.0124}   \\
  
\hline

\end{tabular}
\caption{\textbf{Quantitative evaluation on NeuMan~\cite{jiang2022neuman} dataset.}}
\label{tab:neuman}
\end{table}
}

\newcommand{\tabledyn}{
\begin{table}[t]
\small
\centering
\begin{tabular}{lccc}

\hline  Method & PSNR $\uparrow $ &  SSIM $\uparrow$  & LPIPS $\downarrow $\\ 
\hline
HumanNeRF~\cite{weng2022humannerf}  & 24.39 & 0.9349 &  0.0349\\ 
InstantAvatar~\cite{jiang2023instantavatar} & 20.31  & 0.9183  & 0.1046 \\ 

\hline
Baseline &  23.55 & 0.9340 & 0.0354 \\
Baseline + Opt. & 26.16  & 0.9522 &  0.0237 \\
Baseline + Opt. + Dyn. & \textbf{28.58}  & \textbf{0.9616} & \textbf{0.0169}   \\
\hline

\end{tabular}
\caption{\textbf{Quantitative evaluation on DynVideo dataset.} }
\label{tab:dynvideo}
\end{table}
}

%% file: sections/5_conclusion.tex
\section{Conclusion and Discussion}
\label{sec:conclusion}
We introduce GaussianAvatar, a human avatar reconstruction method based on the proposed animatable 3D Gaussians from monocular videos.
For dynamic human appearance modeling, we leverage a dynamic appearance network along with an optimizable feature tensor to enhance the representation with dynamic properties.
Besides, we implement a joint motion and appearance optimization scheme to rectify estimated motion and enhance the overall reconstruction quality.
Our method shows the capability to reconstruct avatars with dynamic appearances, enabling realistic animation while maintaining real-time rendering speed.
%
%
%
%


\noindent \textbf{Limitation.}
Similar to~\cite{jiang2023instantavatar, weng2022humannerf, yu2023monohuman}, our method may generate artifacts due to inaccurate foreground segmentations in videos and encounter challenges in modeling loose outfits such as dresses.

\noindent \textbf{Potential Social Impact.}
Given our method's capability to reconstruct a realistic personalized character from a monocular video, it is imperative to exercise caution and consider the potential for technology misuse.
\noindent \textbf{Acknowledgement.}
This work was supported by the NSFC project (Nos. 62272134, 62236003, 62072141, 62301298, and 62125107), Shenzhen College Stability Support Plan (Grant No. GXWD20220817144428005), and the Major Key Project of PCL (PCL2023A10-2).

%% file: sections/supp_no_re.tex
\clearpage
\setcounter{page}{1}
\maketitlesupplementary
In the supplementary material, we begin by presenting the implementation details of our method in Sec.~\ref{sec:implementation}.
Following that, we provide information on the proposed dataset in Sec.~\ref{sec:dataset}, conduct the training \& running time comparison in Sec.~\ref{sec:run-time}, and demonstrate the motion optimization comparison in Sec.~\ref{sec:motion}.
Finally, we showcase challenging cases in Sec.~\ref{sec:cases} and present hand animation results in Sec.~\ref{sec:results}.
\section{Implementation Details}
\label{sec:implementation}

\subsection{Model Architecture}
\label{sec:model}
We first estimate the SMPL model for all videos in three datasets.
The input to the pose encoder is the UV map of the SMPL model, which has a resolution of 128 $\times$ 128 $\times$ 3.
We adopt a standard U-Net architecture as the pose encoder, comprising five blocks of [Conv2d, BatchNorm, LeakyReLU], followed by five blocks of [ReLU, ConvTranspose2d, BatchNorm].
Note that we omit the BatchNorm in the final block.

The optimizable feature tensor has the same resolution as the output of the pose encoder, which is 128 $\times$ 128 $\times$ 64.
During the first training stage, we train it using an auto-decoding approach.
Subsequently, the output of the pose encoder is integrated into the optimized feature tensor before being input to the Gaussian parameter decoder. 
To achieve finer details, we conduct a $4\times$ upsampling of the combined feature tensor, resulting in a dimension of 512 $\times$ 512 $\times$ 64.
The resulting output of 3D Gaussians consists of nearly 200,000 points.
The Gaussian parameter decoder comprises an 8-layer Multi-Layer Perceptron (MLP) followed by three prediction heads.
The dimensions of the intermediate layers of the MLP are (128, 128, 128, 256, 128, 128, 128, 64), incorporating a skip connection from the input to the 4th layer.
Each prediction head consists of a 2-layer MLP designed to predict offsets $\Delta \hat{\mathbf{x}}$, colors $\hat{\mathbf{c}}$, and scales $\hat{s}$, respectively.

\subsection{Training}
\label{sec:training}
We first train the optimizable feature tensor and the Gaussian parameter decoder concurrently with motion optimization.
During this stage, we employ the Adam optimizer with specific learning rates: $3.0 \times 10^{-3}$ for the Gaussian parameter decoder, $5.0 \times 10^{-4}$ for the optimizable feature tensor, and $5.0 \times 10^{-3}$ for motion optimization.
We train them for a duration of 200 epochs.
Following this, we generate UV positional maps of SMPL models corrected by optimized motions.
After the first stage of training, we suspend the training of the optimized feature tensor and combine it with the output of the pose encoder.
We proceed to train the pose encoder and fine-tune the Gaussian parameter decoder for an additional 200 epochs.

\section{Dataset Details}
\label{sec:dataset}
We take the same settings in NeuMan for partitioning the proposed DynVideo dataset.
The dataset details are as shown in Table~\ref{tab:dynvideo}.

\begin{table}[t]
\small
\centering
\begin{tabular}{lcccc}

\hline Sequence & Total & Train & Validation & Test \\ 
\hline
male-1 & 978 & 782 & 98 & 98 \\ 
female-1 & 972 & 778 & 97 & 97 \\ 
  
\hline

\end{tabular}
\caption{\textbf{Data distribution.} Number of frames in each sequence used for training, validation, and testing.}
\label{tab:dynvideo}
\end{table}

\begin{table}[t]
\small
\centering
\begin{tabular}{lccc}

\hline  Methods & HumanNeRF & InstantAvatar  & Ours \\ 
\hline
Training time & $\sim$ 13 h & $\sim$ \textbf{1 min}  & $\sim$ 30 min  \\ 
Running time  & 0.22 FPS & 3.87 FPS & \textbf{35 FPS} \\ 
  
\hline
\end{tabular}
\caption{\textbf{Training and running time comparisons.} }
\label{tab:runtime}
\end{table}



    


\section{Training and Running Time Comparison}
\label{sec:run-time}
Here we compare the inference speed of GaussianAvatar with two NeRF-based methods, HumanNeRF and InstantAvatar.
As shown in Table~\ref{tab:runtime}, we measure the training and running time in the People-Snapshot dataset.

\begin{figure}[t]
\includegraphics[width=\linewidth]{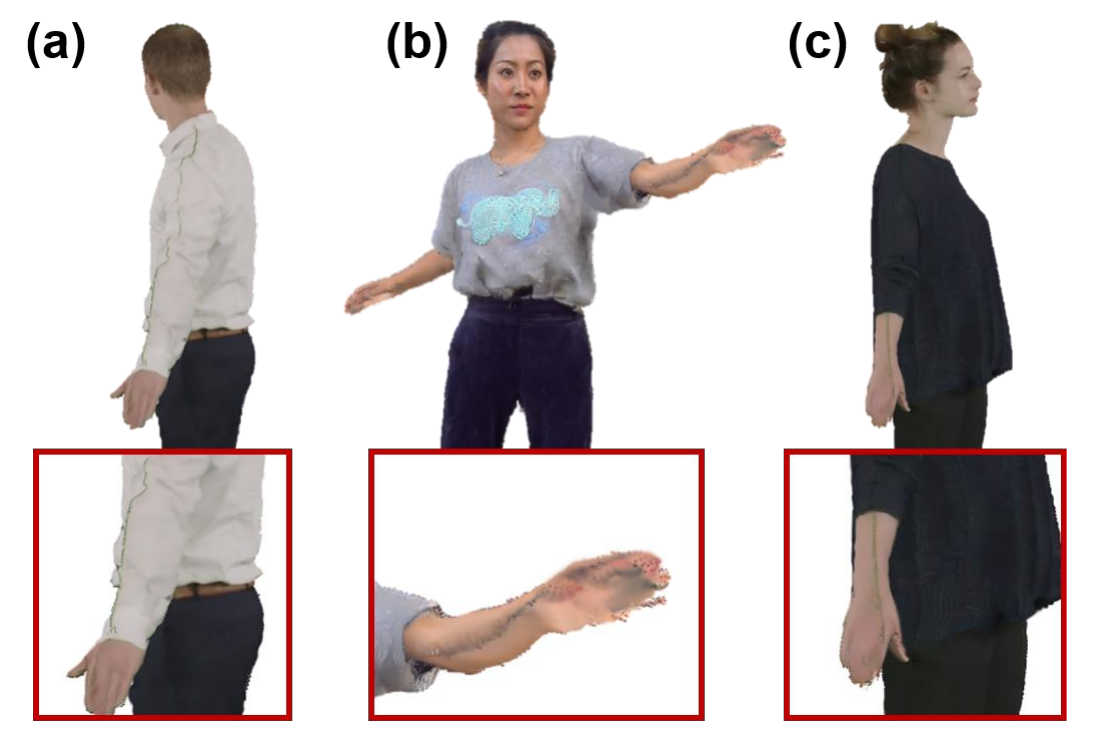}
\caption{\textbf{Results of inaccurate segmentation.} We showcase the artifacts resulting from the inaccurate segmentation boundary.}
\label{fig:seg}
\end{figure}

\begin{figure}[t]
\includegraphics[width=\linewidth]{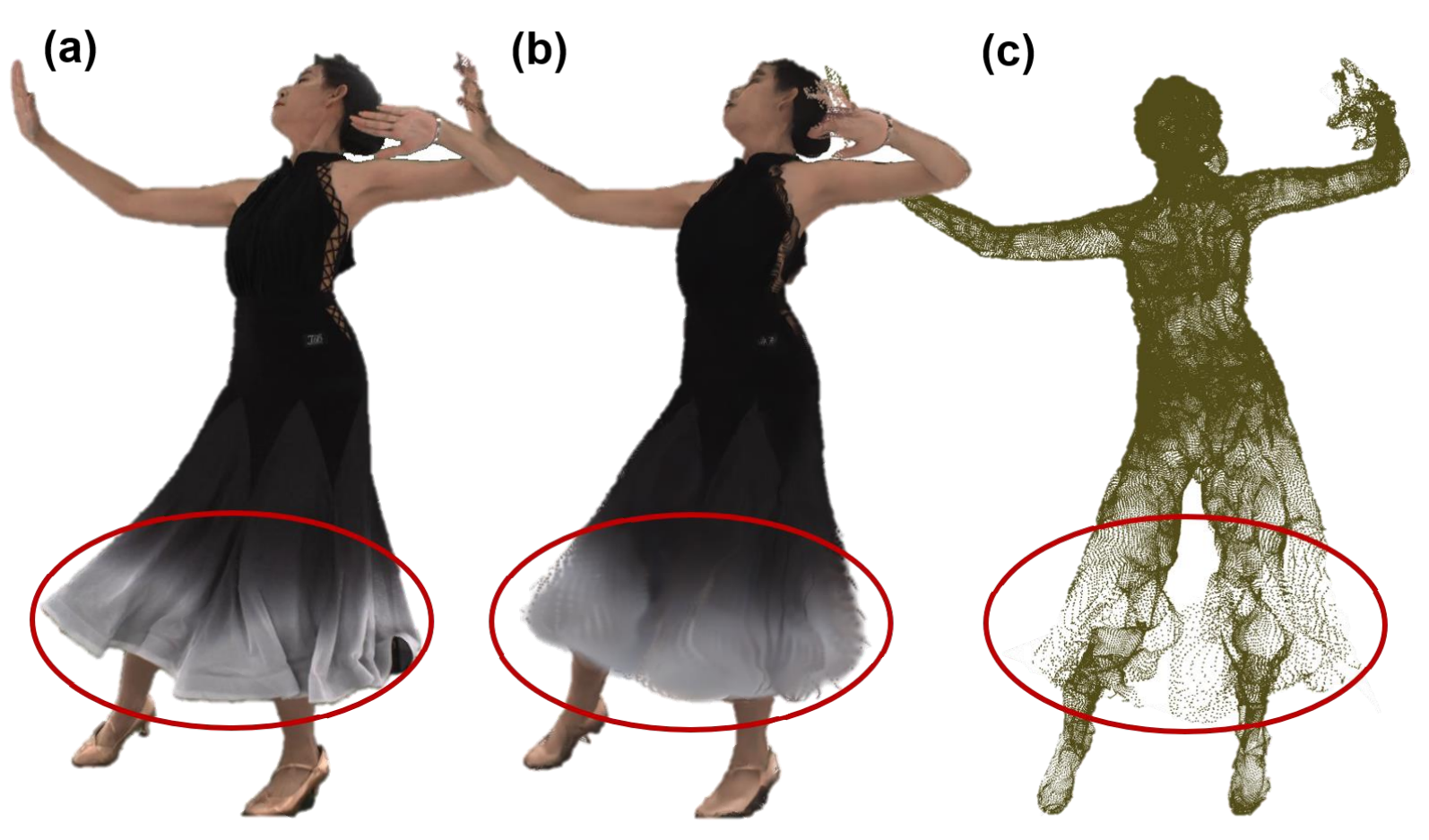}
\caption{\textbf{Results of loose clothing.} (a) is the ground truth, (b) and (c) are the rendered image and Gaussian points.}
\label{fig:dress}
\end{figure}

\section{Motion Optimization Comparison}
\label{sec:motion}
\begin{figure}[t]
\includegraphics[width=\linewidth]{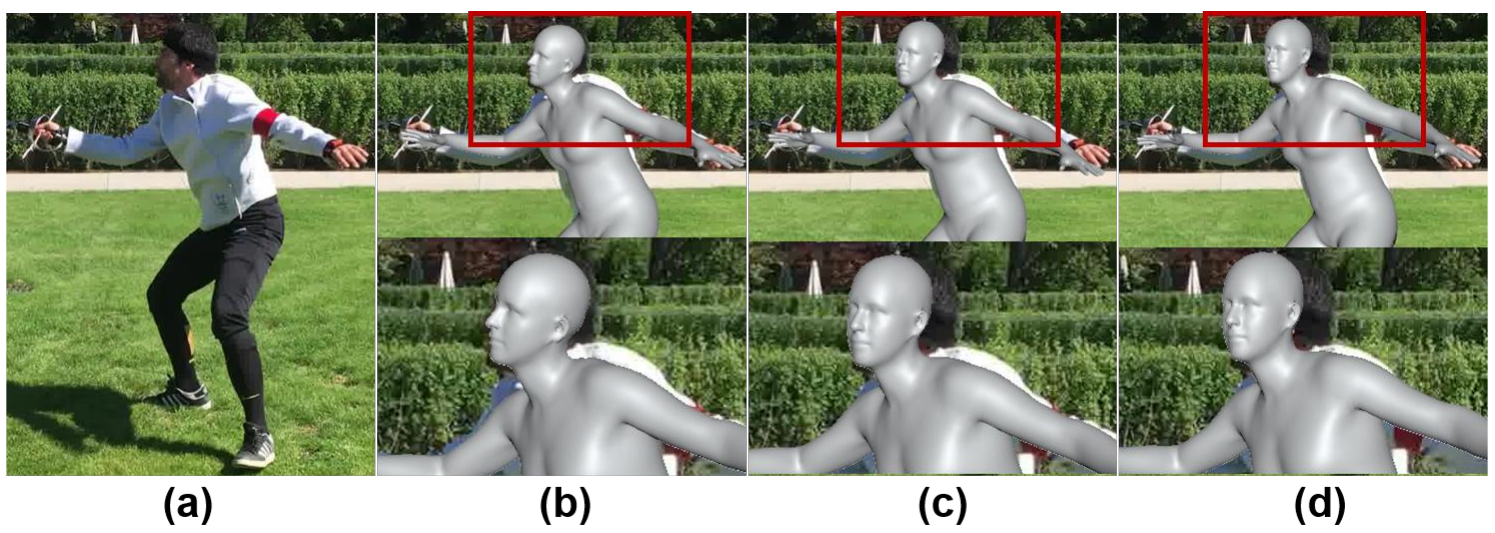}

\caption{\textbf{Results of motion optimization comparison.} (a) Original image
, (b) our optimized SMPL, (c) refined SMPL by InstantAvatar, (d) initial SMPL.}
\label{fig:supp_opt}
\end{figure}

We directly evaluate the pose refinement of GaussianAvatar and the SOTA InstantAvatar on two sequences in the 3DPW dataset and one sequence in the DNA-Rendering dataset.
Both Table~\ref{tab:opt} and Fig.~\ref{fig:supp_opt} show that our GaussianAvatar outperforms InstantAvatar in pose refinement.

\begin{table}[t]
\small
\centering
\begin{tabular}{lccc}

\hline  Methods & Initial motion & InstantAvatar  & Ours \\ 
\hline
P-MPJPE & 71.95 & 70.87 & \textbf{64.94}  \\ 
  
\hline
\end{tabular}
\caption{\textbf{Motion optimization comparison.}}
\label{tab:opt}
\end{table}

\section{Challenging Cases}
\label{sec:cases}
As discussed in the final section of the main paper, a major limitation of our approach is attributed to the inaccuracies in foreground segmentation in videos.
As shown in Fig.~\ref{fig:seg}, the inaccuracies in the foreground segmentation boundary may lead to our method predicting a black line on the surface.
Automatic segmentation tools do not always yield satisfactory segmentation results.
Manual operations on these segmentations are time-consuming and inefficient.
We believe that addressing this issue can be achieved by incorporating a scene model, akin to approaches such as NeuMan and Vid2Avatar, which can contribute to more accurate segmentation. 
We leave this for future work.
%

%
Besides, modeling the dynamic appearance of dresses remains challenging. 
As shown in Fig.~\ref{fig:dress}, our method produces a blurred clothing appearance and fails to reconstruct complete point clouds.
The primary challenge stems from the derived skinning weights from the SMPL model.
Employing these skinning weights to model dresses may lead to artifacts when generalized to new poses.
The prospect of predicting specific skinning weights for each subject is promising.
However, this data-driven approach necessitates specific data sources.
We intend to collect this kind of data in future efforts.



\section{Hand Animation}
\label{sec:results}
\begin{figure}[t]
\includegraphics[width=\linewidth]{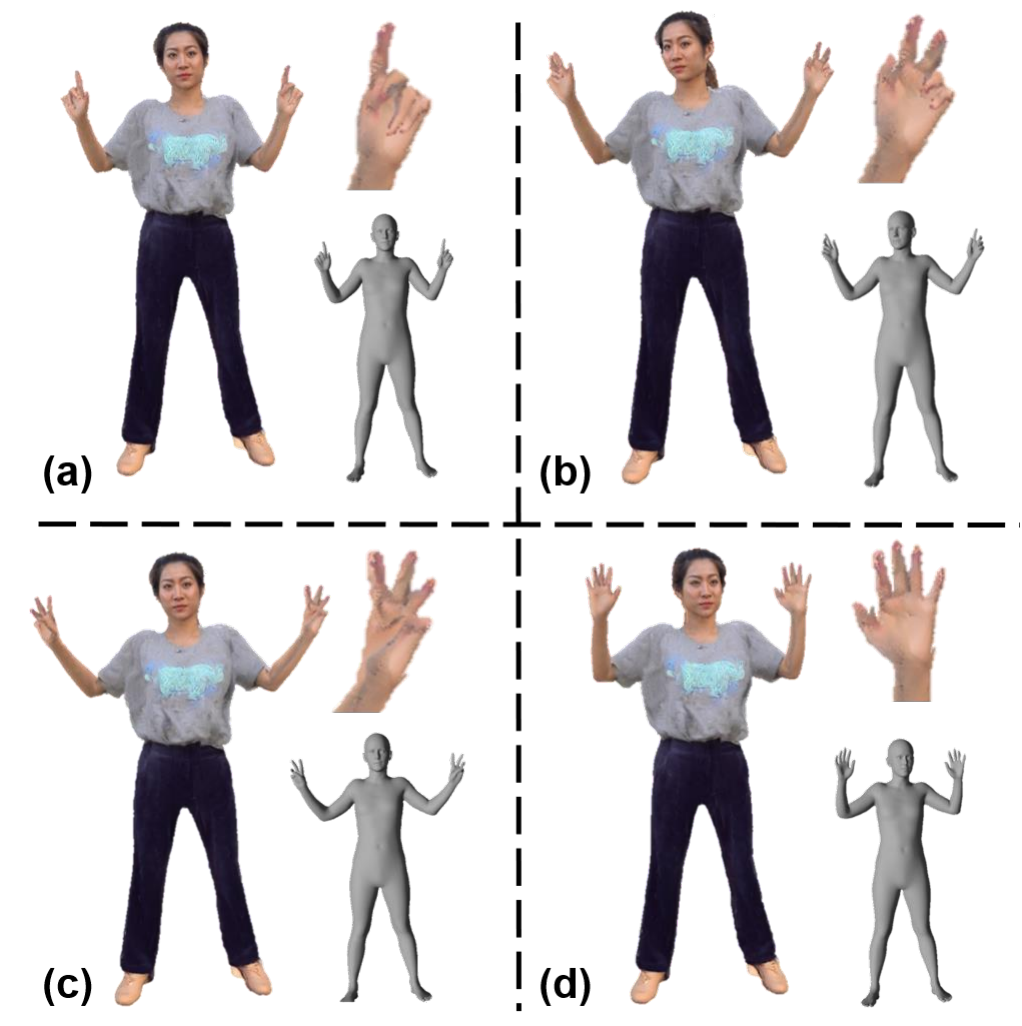}

\caption{\textbf{Results of hand animation.} (a-d) Left: reposed image, bottom right: reference pose.}
\label{fig:hand}
\end{figure}

%
%
%
We observe that our method can be readily extended to hand animation. 
To showcase its effectiveness in this context, we estimate the underlying SMPL-X model to fit a sequence from the DynVideo dataset. 
As depicted in Fig.~\ref{fig:hand}, our method demonstrates the capability to generate plausible hand animation without the need for specific design considerations. 
The prospect of extending our work to encompass full-body avatars is promising, and we defer this to future investigations.

